\documentclass[a4paper]{article}
\usepackage{bm}
\usepackage{tabularx} 

\newcommand{\vect}[1]{\bm{#1}}
\newcommand{\matr}[1]{\bm{#1}}

\newcommand{\va}[0]{\vect{a}}

\newcommand{\vc}[0]{\vect{c}}
\newcommand{\ve}[0]{\vect{e}}
\newcommand{\valpha}[0]{\vect{\alpha}}

\newcommand{\vh}[0]{\vect{h}}
\newcommand{\vv}[0]{\vect{v}}

\newcommand{\vz}[0]{\vect{z}}

\newcommand{\vl}[0]{\vect{l}}
\newcommand{\vs}[0]{\vect{s}}

\newcommand{\vi}[0]{\vect{i}}
\newcommand{\vo}[0]{\vect{o}}

\newcommand{\vr}[0]{\vect{r}}

\newcommand{\mW}[0]{\matr{W}}

\newcommand{\mE}[0]{\matr{E}}

\newcommand{\mU}[0]{\matr{U}}

\newcommand{\mL}{\matr{L}}
\usepackage{color}

\definecolor{olive}{RGB}{0,184,136}

\usepackage{INTERSPEECH_v2}

\title{Visually Grounded Word Embeddings and Richer Visual Features for Improving Multimodal Neural Machine Translation}
\name{Jean-Benoit Delbrouck$^1$, St\'ephane Dupont$^1$, Omar Seddati$^1$}
\address{
  $^1$TCTS Lab, University of Mons, Belgium}
\email{ \{Jean-Benoit.DELBROUCK, Stephane.DUPONT, Omar.SEDDATI\}@umons.ac.be}

\begin{document}

\maketitle
\begin{abstract}
In Multimodal Neural Machine Translation (MNMT), a neural model generates a translated sentence that describes an image, given the image itself and one source descriptions in English. This is considered as the multimodal image caption translation task. The images are processed with Convolutional Neural Network (CNN) to extract visual features exploitable by the translation model. So far, the CNNs used are pre-trained on object detection and localization task. We hypothesize that richer architecture, such as dense captioning models, may be more suitable for MNMT and could lead to improved translations. We extend this intuition to the word-embeddings, where we compute both linguistic and visual representation for our corpus vocabulary. We combine and compare different configurations and show state-of-the-art results according to previous work. 

\end{abstract}
\noindent\textbf{Index Terms}: neural machine translation, multimodal, attention mechanism, image features, visual embeddings, grounding

\section{Introduction}

Neural networks have shown great performance on the Machine Translation (MT) task. The encoder-decoder framework \cite{SutskeverVL14} has been since widely adopted. An attention mechanism has been introduced by Bahdanau et al. \cite{BahdanauCB14} to learn to focus on different parts of the input sentence while decoding. Other modalities, like images, can make use of such attention mechanisms. A previous work \cite{icml2015_xuc15} has shown they are able to learn to attend to the salient parts of an image when generating a text captions. \\

Integrating multimodal information efficiently still remains a challenge. It requires combining diverse modality vector representations. A few attempts  \cite{Calixto2016DCUUvAMM, libovicky-EtAl:2016:WMT, caglayan2016does,Huang2016AttentionbasedMN,SHEFShah} have been made during the WMT 2016 Multimodal Machine Translation evaluation campaign. These initial efforts have not convincingly demonstrated that visual context can improve translation quality. Meanwhile, few improvements have been made, \cite{CalixtoLC17b} proposed a doubly-attentive decoder that outperformed all previous baselines with less data and without re-scoring, \cite{delbrouck-dupont:2017:EMNLP2017} tried multiple attention models and image attention optimizations such as the gating \cite{icml2015_xuc15} and pre-attention \cite{DBLP:journals/corr/DelbrouckD17} mechanism. Recently, \cite{DesmondImagination} introduced a model where visually grounded representations are learned.\\

In this paper, our aim is to propose a first empirical investigation on improving MNMT by using improved visual and word representations. More specifically, we believe that visual and word representations obtained through models pre-trained on large data sets should bring further improvement. Most importantly, we want to leverage models that provide a closer link between image understanding and language understanding. For extracting image modality vector representations, we will make use of a model trained on a dense captioning task, namely DenseCap \cite{densecap}. Compared to models trained on object recognition tasks (such as ImageNet \cite{Russakovsky:2015:ImageNet}, as used in previous MNMT proposal), the hope is that the representation contains richer information, also encoding object attributes and important relationship for linguistic description of the images. For extracting the vector representations of the word modality, we will make use of word vectors obtained from large scale text corpora, but we will also use visual representations of the referents of those words, using a recent paradigm of "imagined" visual representations of those words. The hope is that these visually grounded word representation will facilitate the integration of both modalities during the decoding process, hence improving translation results.\\ 

Our paper is structured as follows. In section \ref{model}, we briefly describe our NMT model as well as the conditional GRU activation used in the decoder. We also explain how multi-modalities can be implemented within this framework. In the following section \ref{vembeddings} and \ref{vfeatures}, we detail the process of our visual embeddings and features creation. Finally, we report and analyze our results in section \ref{resultsconcl}.
    
\section{Model} \label{model}
\subsection{Text-based NMT}

We describe the attention-based NMT model introduced by \cite{BahdanauCB14} in this section. Given a source sentence $X = (x_1, x_2, \hdots , x_M)$, the neural network directly models the conditional probability $p(Y|X)$ of its translation $Y = (y_1, y_2, \hdots, y_N)$. The network consists of one encoder and one decoder with one attention mechanism. 
Each source word $x_i$ and target word $y_i$ are a column index of the embedding matrices $\mE_X$ and $\mE_Y$. The encoder is a bi-directional RNN with Gated Recurrent Unit (GRU) layers \cite{ChungGCB14,cho-al-emnlp14}, where a forward RNN $\overrightarrow{\Psi}_\text{enc}$ reads the input sequence as it is ordered (from $x_1$ to $x_M$) and calculates a sequence of forward hidden states $(\overrightarrow{\vh}_1, \overrightarrow{\vh}_2, \hdots, \overrightarrow{\vh}_M)$. A backward RNN $\overleftarrow{\Psi}_\text{enc}$ reads the sequence in the reverse order (from $x_M$ to $x_1$), resulting in a sequence of backward hidden states $(\overleftarrow{\vh}_M, \overleftarrow{\vh}_{M-1}, \hdots, \overleftarrow{\vh}_1)$. We obtain an annotation for each word $x_i$ by concatenating the forward and backward hidden state $\vh_t = [\overrightarrow{\vh}_t;\overleftarrow{\vh}_t]$. Each annotation $\vh_t$ contains the summaries of both the preceding words and the following words. The representation $C$ for each source sentence is the set of annotations $C = (\vh_1, \vh_2, \hdots, \vh_M)$.\\

The decoder is an RNN that uses a conditional GRU\footnote{https://github.com/nyu-dl/dl4mt-tutorial/blob/master/docs/cgru.pdf} (cGRU) with an attention mechanism to generate a translated word $y_t$. More precisely, the conditional GRU has three main components computed at each time step $t$ of the decoder:

\begin{itemize}
\item REC1 computes a hidden state proposal $\vs_t^{\prime}$
based on the previous hidden state $\vs_{t-1}$ and the previously emitted word $y_{t-1}$;
\item $f_{\text{att}}$ \footnote{called ATT in the aforementioned paper} is an attention mechanism over the hidden states of the encoder and computes a time-dependent context vector $\vc_t$ using the annotation set $C$ and the hidden state proposal $\vs_t^{\prime}$;
\item REC2 computes the final hidden state $\vs_t$ using the hidden state proposal $\vs_t$ and the context vector $\vc_t$;
\end{itemize}
    
Both $\vs_{t}$ and $\vc_t$ are further used to decode the next symbol. We use a deep output layer \cite{Pascanu2014} to compute a vocabulary-sized vector :
	\begin{equation}
	\vo_t=\mL_o \tanh (\mL_ss_t + \mL_c\vc_t + \mL_w\mE_Y[y_{t-1}]) \label{deepout}
	\end{equation}
	where $\mL_o$, $\mL_s$, $\mL_c$, $\mL_w$ are model parameters. We can parameterize the probability of decoding each word $y_t$ as:
	\begin{equation}
	p(y_t | y_{t-1}, \vs_t, \vc_t) = \text{Softmax}(\vo_t)
	\end{equation}
	The initial state of the decoder $\vs_0$ at time-step $t=0$ is initialized by the following equation : 
	\begin{equation}
	\vs_0 = f_{\text{init}}(\vh_M) \label{initdec}
	\end{equation}
	where $f_{\text{init}}$ is a feed-forward network with one hidden layer. \\

We use the soft attention mechanism for the $f_{\text{att}}$ component. Soft attention has firstly been used for syntactic constituency parsing by \cite{NIPS2015Vinyals} but has been widely used for translation tasks ever since. The idea of the soft attentional model is to consider all the annotations when deriving the context vector $\vc_t$.  It consists of a single feed-forward network used to compute an expected alignment $\ve_{t}$ between text annotation $\vh_i$ and the target word to be emitted at the current time step $t$. The inputs are the annotations and the intermediate representation of REC1 $\vs_t^{\prime}$:

\begin{equation}
	\ve_{t,l} = \vv^T \tanh(\mU_a \vs_t^{\prime} + \mW_a \va_l ) \label{eqenergy}
	\end{equation}
    
	\begin{align}
	\valpha_{t,i} =& \dfrac{\exp(\ve_{t,i})}{\sum\nolimits_{j=1}^L\exp(\ve_{t,j})} \label{eqalpha}
    \end{align}

where $\valpha_{t,i}$ is the normalized alignment matrix between each source annotation vector $h_i$ and the word $y_t$ to be emitted at time step $t$. In the above expressions, $\vv^T$, $\mU_a$ and $\mW_a$ are trained parameters. Finally, the modality time-dependent context vector $\vc_t$ is computed as a weighted sum over the annotation vectors (equation \ref{softsum}).
	\begin{align}
	\vc_{t} =& \sum\limits_{i=1}^L \valpha_{t,i}\va_i \label{softsum}
	\end{align}
 
\subsection{Multimodal NMT (MNMT)}
In multimodal NMT, the second modality is usually an image, for which feature maps are computed using a Convolutional Neural Network (CNN). The annotations $\va_1,\va_2, \hdots, \va_L$ are spatial features  (i.e. each annotation represents features for a specific region in the image). More formally, given a set of image modality annotations $I=(\va_1,\va_2, \hdots, \va_L)$, we compute a an image context vector $\vi_t$ based on the same intermediate hidden state proposal: 
	\begin{equation}
	\vi_t = f_{\text{att}}^\prime \left(  \text{I}, \vs_t^{\prime}  \right) 
	\end{equation}	
	This new time-dependent context vector is an additional input to a modified version of REC2 which now computes the
	final hidden state $\vs_t$ using the intermediate hidden state proposal $\vs_t^{\prime}$ and both time-dependent context vectors $\vc_t$ (for text) and $\vi_t$ (for image). In addition, $\vi_t$ is weighted with the gating scalar mechanism as seen in \cite{icml2015_xuc15}\\	
	\begin{align}
	\vz_t =& \sigma \left( \mW_z \vc_t + \mW_z \vi_t + \mU_z \vs_t^{\prime} \right)  \nonumber \\
	\vr_t =& \sigma \left( \mW_r \vc_t + \mW_r \vi_t + \mU_r \vs_t^{\prime} \right) \nonumber  \\
	\underline{\vs}_t =& \text{tanh} \left(  \mW \vc_t  + \mW \vi_t  + \vr_t \odot (\mU \vs_t^{\prime} )  \right) \nonumber \\       
	\vs_t =& (1 - \vz_t) \odot \underline{\vs}_t + \vz_t \odot \vs_t^{\prime}
	\end{align}	
	The probabilities for the next target word (from equation \ref{deepout}) also takes into account the new context vector $\vi_t$:
	\begin{equation}
	\mL_o \tanh (\mL_s\vs_t + \mL_c\vc_t +  \mL_i\vi_t + \mL_w\mE_Y[y_{t-1}])
	\end{equation}
	where $\mL_i$ is a new trainable parameter. 

\section{Improved Word Embeddings} \label{vembeddings}
In previous works on MNMT, word embeddings are usually trained along with the model. Both matrices $\mE_X$ and $\mE_Y$ are considered trained model parameters. This approach does not allow to exploit large scale text corpora that can be available in the source language and that could be leveraged to obtain useful distributed semantic representations of the words (such as Word2Vec \cite{NIPS2013_W2V}, or Glove \cite{pennington2014glove}). Here, we will make use of Glove to build a multimodal representation, textual and visual, for our whole source vocabulary. To do so, we try out an effective method that learns a language-to-vision mapping as described in \cite{collell2017imagined}. The learned model outputs visual predictions of a word given its semantic representation.  

\subsection{Language-to-vision mapping}

Concretely, we consider two embedding spaces: a linguistic space $\mathcal{L} \subset \mathbb{R}^{d_l}$ and a visual space $\mathcal{V} \subset \mathbb{R}^{d_v}$ where $d_l$ and $d_v$ are the sizes of the text and visual representations respectively. For a given dataset of words $\mathcal{W} = \{w_1, w_2, \hdots, w_N\}$, each word $w_i$ has a linguistic representation $\vl_{w_i}  \in \mathcal{L}$ and a visual representation $\vv_{w_i}  \in \mathcal{V}$. The aim is to learn a mapping
function $f :  \mathcal{L} \rightarrow \mathcal{V}$ such that the prediction (or imagined representations) $f(\vl_{w_i})$ is close to the actual visual vector $\vv_{w_i}$. A training example is thus a pair $\{\vl_{w_i};\vv_{w_i}\}$ and the dataset is composed of $N$ examples.  \\

\subsection{Visually Grounded Word Embeddings}\label{groundedwe}

ImageNet is used as source of visual information. ImageNet covers a total of 21,841 WordNet synsets and has 14,197,122 images. For the experiment, only synsets with more than 50 images are kept, and an upper bound of 500 images per synset is used to reduce computation time.  With this restriction, 9,251 unique words are covered. The training set is composed of $N$ = 9,251 examples. For each of these words, we use a pre-trained VGG-m-128 CNN model \cite{Simonyan14c} to extract visual features from each image. We take the 128-dimensional activation of the last layer. The visual representation $\vv_{w_i}$ is computed as an element-wise averaging of the features vectors from different images picturing the object the word refers to. As previously mentioned, the textual representation $\vl_{w_i}$ is obtained with the word embeddings algorithm Glove. We use the pre-trained model on the Common Crawl corpus consisting of 840B tokens and a 2.2M words. The mapping function $f$ consists of a simple perceptron composed of a $d_l$ dimensional input layer and a linear output layer with $d_v$ units.\\

\subsection{Imagined multimodal embeddings}
We use the pre-trained model made available by \cite{collell2017imagined}. Their training is done with a mean squared error (MSE) loss function and stochastic gradient descent optimizer. A learning rate of 0.1 and dropout rate of 0.1 is chosen, running for 175 epochs. GloVe vectors are of size $d_l$ = 300 and a low-dimensional $d_v$ = 128 is picked to reduce the number of parameters and thus the risk of overfitting. The multimodal representation of word is built by concatenating the $\ell$2-normalized imagined representations $f(\vl_{w_i})$ with the textual representations $\vl_{w_i}$.  Hence, the multimodal representation is of size 428. We compute it using our model for every word in our source vocabulary.

	\begin{table*} 
    		\caption{Results on the test triples of the Multi30K dataset. "Along" means embeddings are initialized with a Gaussian and trained along with the model. "Fixed" means that the embeddings are frozen during the whole training. Embeddings size are between brackets at the end of each model description. Each score is compared with the model B1.}
		\centering
		\begin{tabular}{lcccccc}
			\multicolumn{1}{c}{\bf Model}  &\multicolumn{6}{c}{\bf 				Test Scores}
			\\ \hline \\
			&BLEU$\uparrow$& & METEOR$\uparrow$& &TER$\downarrow$ & \\ 
            \textbf{Previous work} \\
            
            \cite{CalixtoLC17b} ResNet-50 + along (620)       &36.50& &55.0& &43.7& \\
            \cite{DesmondImagination} GoogLeNet v3 +  along (620)      &36.8 $\pm$ 0.8& &55.8 $\pm$ 0.4& &-&\\
            \cite{DesmondImagination} GoogLeNet v3 +  along (620) + COCO + NC  &37.8 $\pm$ 0.7& &57.1 $\pm$ 0.2& &-&
                        \\
            & \\
            \hline \\
            \textbf{Baseline} \\            
			(B1) ResNet-50 + along (428)            &36.27& &53.9& & 43.6& \\
            (B2) DenseCap  + along (428)          &37.78&\textcolor{olive}{$\uparrow$ +1.51}& 54.6&\textcolor{olive}{$\uparrow$ +0.7}& 42.3&\textcolor{olive}{$\downarrow$ -0.3}  \\
            & \\
            
            \textbf{Linguistic word embeddings} \\
            (L1) ResNet-50 + GloVe (300)        &37.40& \textcolor{olive}{$\uparrow$ +1.13}&55.0&\textcolor{olive}{$\uparrow$ +1.1}&42.1&\textcolor{olive}{$\downarrow$ -0.5}  \\
            (L2) DenseCap + GloVe (300) &37.95&\textcolor{olive}{$\uparrow$ +1.68}&55.7&\textcolor{olive}{$\uparrow$ +1.8}&42.1&\textcolor{olive}{$\downarrow$ -0.5} \\
			& \\          
           
           \textbf{Imagined multimodal word embeddings} \\			
			(M1) ResNet-50 + GloVe + Visual (428, fixed)      &35.52&\textcolor{red}{$\downarrow$ -0.75}&53.7&\textcolor{red}{$\downarrow$ -0.2}&43.3&\textcolor{red}{$\uparrow$ +0.7} \\
			(M2) ResNet-50 + GloVe + Visual (428)&37.51&\textcolor{olive}{$\uparrow$ +1.24}&55.2&\textcolor{olive}{$\uparrow$ +1.3}&42.1&\textcolor{olive}{$\downarrow$ -0.5} \\

            (M3) DenseCap + GloVe + Visual (428)&38.20&\textcolor{olive}{$\uparrow$ +1.93}&55.7&\textcolor{olive}{$\uparrow$ +1.8} &41.9&\textcolor{olive}{$\downarrow$ -0.7} \\  
         
		\end{tabular}    
        \label{table1}
	\end{table*}%

\section{Visual Features} \label{vfeatures}
\begin{figure}[!ht]
		\centering
		\includegraphics[scale=0.39]{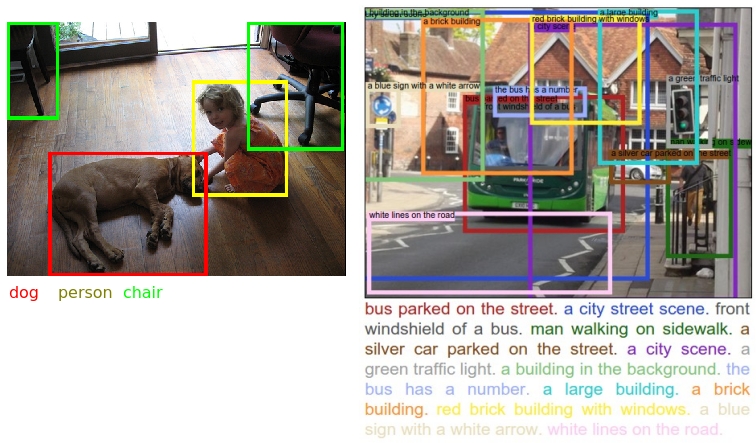}
		\caption{Left: Object localization and detection. Right: Dense
        captioning}
                \vspace{-12pt}
     \label{detectioncaptioning}
	\end{figure}

So far in MNMT, the spatial visual features $\va_1,\va_2, \hdots, \va_L$ of the images are extracted with a 16 or 19-layers version of VGGNet, or a Deep Residual Network \cite{He_2016_CVPR} pre-trained on ImageNet for an object detection and localization task. As mentioned in \cite{JBempiricalEMNLP}, such features may not be suited for the translation of complex captions, which involves objects but also their attributes and relationships (as shown in Figure \ref{detectioncaptioning}). In this work, we focus on using features extracted on a model pre-rained for a dense captioning task, namely DenseCap pre-trained on Visual Genome \cite{krishnavisualgenome}. Its architecture is composed of a Convolutional Network, a novel dense localization layer, and Recurrent Neural Network language model that generates the label sequences. The CNN is pretrained on ImageNet and fine-tuned during the training (except for the first four convolutional layers). We extract the features at the last convolutional layer. Due to the high sparsity of these features, we apply an $\ell_2$-normalization.\\
In this experiment, we compare two image annotations used by our decoder: features of size $14 \times 14 \times 1024$ extracted from a ResNet-50 pre-trained on ImageNet (at its res4f layer) and $14 \times 14 \times 512$ from DenseCap as described above.

\section{Dataset and model settings} \label{settings}

For this experiments on Multimodal Machine Translation, we used the Multi30K dataset \cite{elliott-EtAl:2016:VL16} which is an extended version of the Flickr30K Entities. For each image, one of the English descriptions was selected and manually translated into German by a professional translator. As training and development data, 29,000 and 1,014 triples are used respectively. A test set of size 1000 is used for metrics evaluation. \\

	All our models are build on top of the nematus framework \cite{nematus}. The encoder is a bidirectional RNN with GRU, one 1024D single-layer forward and one 1024D single-layer backward RNN. Non-recurrent matrices are initialized by sampling from a Gaussian $\mathcal{N}(0, 0.01^2)$, recurrent matrices are random orthogonal and bias vectors are all initialized to zero. The word embeddings matrices $\mE_X$ and $\mE_Y$ are either trained along the model and initialized accordingly or pre-trained as described in section \ref{vembeddings}. Embeddings  size depends on the experiment and are explicitly mentioned in the score tabular (Table \ref{table1}) for every model. We apply dropout with a probability of 0.3 on the embeddings, on the hidden states in the bidirectional RNN in the encoder as well as in the decoder. In the decoder, we also apply dropout on the text annotations $\vh_i$, the image features $\va_i$, on both modality context vector and on all components of the deep output layer before the readout operation. Dropout is applied using one same mask in all time steps \cite{Gal2016Theoretically}. \\ 
    
	We normalize and tokenize English and German descriptions using the Moses tokenizer scripts \cite{Koehn:2007}. We use the byte pair encoding algorithm on the target train set to convert space-separated tokens into subwords \cite{sennrich2016subword}, reducing the German vocabulary to 14957 words. Our source vocabulary, in English, is of size 11187. The visual features of the Flickr30K images are extracted with DenseCap or ResNet-50 in our experiments (as described in section \ref{vfeatures}).  \\ 
    
	All variants of our model were trained with ADADELTA \cite{ADADELTAZeiler}, with mini-batches of 40 examples. We apply early stopping for model selection based on BLEU4 : training is halted if no improvement on the development set is observed for more than 20 epochs. We use the metrics BLEU4 \cite{Papineni:2002}, METEOR \cite{meteor-wmt:2014} and TER \cite{Snover06astudy} to evaluate the quality of our models' translations.

\section{Results and Future work} \label{resultsconcl}
We report our results in table \ref{table1}. We structure our analysis in three main sections. We start by discussing the effectiveness of the DenseCap features, followed by our impressions about the impact of the multimodal word embeddings. Finally, we qualitatively compare the translations of some of our models to get more concrete and tangible results of our choices. \\

\textbf{Flickr30K Visual Features} \quad We observe a noticeable improvement when using DenseCap visual features instead of ResNet-50. We show an amelioration, between model B1 and B2, of +1.51 BLEU and +0.7 METEOR when embeddings are trained along with the model. With pre-trained GloVe embeddings, between systems L1 and L2, we notice an improvement of +0.55 BLEU and 0.7 METEOR. Lastly, with multimodal embeddings, model M3 scores +0.69 BLEU and +0.5 METEOR over M2. Overall, the results confirm our intuition that rich dense captioning task improves translation's quality, especially on the METEOR metric which proves that the attention model benefits from this change. The use of pre-trained embeddings lowers the gap betweens image features efficiency (ie. the scores difference between M2-M3 and L1-L2 are slightly lower than M1-M2). Moreover, it is interesting to note that the Denscap features (of size $14 \times 14 \times 512$) are twice smaller than the ResNet-50 features ($14 \times 14 \times 1024$). \\

\textbf{Word embeddings} \quad  Obviously, adding GloVe word embeddings helps the model to better translate. Yet, the addition of the visual embeddings to the linguistic embeddings only brings a small improvement: +0.11 BLEU and +0.2 METEOR for ResNet-50 (L1-M2) and +0.35 BLEU and +0.0 METEOR for DenseCap (L2-M3). We can hypothesize two reasons. Firstly, the visual embeddings of size 128 may be too small to be really significant in rather large networks such as these presented and the model in section \ref{vembeddings} might need some changes. Another explanation could be that the visual information used during the training of the mapping function (the function $f$ that outputs our visual embeddings of size 128) are extracted with a VGG-128 network. However, our models are trained using a ResNet-50 or DenseCap network for visual features extraction. It is possible that too much model parameters are requested to do the mapping between the two embedding spaces (the word embeddings and the visual features) and therefore impact the models' translation quality. On a side note, we tried to freeze the loaded embeddings during the training (referred as "fixed" in the score tabular) but lead to unpleasant results of -0.75 BLEU and -0.2 METEOR (B1-M1). We conclude that the models needs to slightly change the representations of the words in order to generate the strongest textual and visual context vectors.\\

\begin{figure}[!ht]
		\centering
		\includegraphics[scale=1.15]{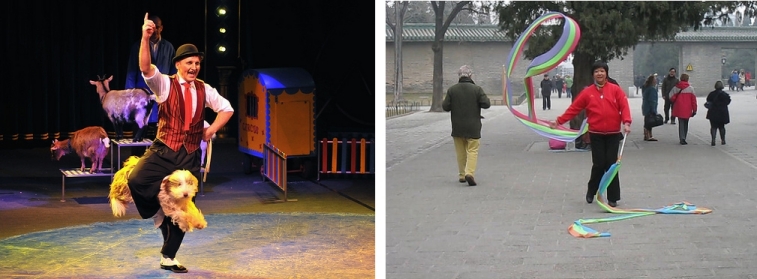}
		\caption{Left: First sentence Right: Second sentence}
         \label{flickrimage}
         \end{figure}

\textbf{Translation comparison} \quad We illustrate some hand-picked examples of significant improvements on the test set we noticed between our models. We start by comparing the two baseline systems B1 and B2. We pick a sentence that involves a positional relationship between a man and a dog (Figure \ref{flickrimage}):

	\begin{table}[h!]
		\centering
		\begin{tabular}{ll}
        	Source:&A man is dancing with a dog between his legs .\\
			Reference:&Ein Mann tanzt mit einem Hund zwischen den \\
            &Beinen .\\
            B1:&Ein Mann tanzt \textcolor{red}{und ein} Hund zwischen \textcolor{red}{seinen} \\
            &Beinen .\\
			B2:&Ein Mann tanzt mit
 einem Hund zwischen den \\
            &Beinen .
		\end{tabular}
	\end{table}
    
The sentence-level BLEU score is of 100 for B2 (DenseCap) and 31.40 for B1 (ResNet-50).
We now compare our two best models M2 and M3 on a more complex and descriptive sentence:

	\begin{table}[h!]
		\centering
		\begin{tabular}{ll}
        	Source:&A person in a red jacket with black pants \\
            &holding rainbow ribbons .\\
			Reference:&Eine Person in einer roten Jacke und schwarzen \\
            &Hosen h{\"a}lt Regenbogenb{\"a}nder .\\
            M2:&Eine Person \textcolor{red}{in roter} Jacke \textcolor{red}{mit} schwarzen \\ 
            &Hosen h{\"a}lt \textcolor{red}{eine Znde in der Hand} .\\
			M3:&Eine Person in einer roten Jacke und schwarzen \\ 
            &Hosen h{\"a}lt \textcolor{red}{einen Zebnde} .
		\end{tabular}
	\end{table}
    
M3 and M2 respectively have a sentence-level BLEU score of 77.19 and 23.66.\\

In future work, it would be interesting to use the same CNN for all the model's components (multimodal word embeddings and visual features). The experiment could be re-attempt with DenseCap on all fronts. Another work could be a more in-depth investigation of the attention model behavior using DenseCap. Its not clear if the improved behavior learned by the model is brought by the word-embeddings (and therefore GloVe and the VGG-128 of section \ref{vembeddings}) or the attention model on the visual features (extracted with DenseCap).

\section{Acknowledgements}
This work was partly supported by the Chist-Era project IGLU with contribution from the Belgian Fonds de la Recherche Scientique (FNRS), contract no. R.50.11.15.F, and by the FSO project VCYCLE with contribution from the Belgian Waloon Region, contract no. 1510501.

\bibliographystyle{IEEEtran}

\bibliography{mybib}


\end{document}